\documentclass[10pt,twocolumn,letterpaper]{article}

\usepackage{cvpr}
\usepackage{times}
\usepackage{epsfig}
\usepackage{graphicx}
\usepackage{amsmath}
\usepackage{amssymb}
\usepackage{bm}
\usepackage{array}
\usepackage{subcaption,graphicx}

\newcommand\blfootnote[1]{%
  \begingroup
  \renewcommand\thefootnote{}\footnote{#1}%
  \addtocounter{footnote}{-1}%
  \endgroup
}

\usepackage[pagebackref=true,breaklinks=true,letterpaper=true,colorlinks,bookmarks=false]{hyperref}

\cvprfinalcopy % *** Uncomment this line for the final submission

\usepackage{adjustbox}

\def\cvprPaperID{6381} % *** Enter the CVPR Paper ID here

\ifcvprfinal\fi

\begin{document}

\definecolor{pierrecolor}{rgb}{0.1,0.9,0.1}
\newcommand{\pierre}[1]{{\textcolor{pierrecolor}{[PM #1]}}}
\newcommand\pma[1] {\pierre{#1}}

\definecolor{stevencolor}{rgb}{0.1,0.1,0.9}
\newcommand{\steven}[1]{{\textcolor{stevencolor}{[SGM #1]}}}
\newcommand\sgm[1] {\steven{#1}}

\definecolor{sarahcolor}{rgb}{0.858,0.188,0.478}
\newcommand{\sarah}[1]{{\textcolor{sarahcolor}{[SP #1]}}}
\newcommand\sap[1] {\sarah{#1}}

\definecolor{gregcolor}{rgb}{0.358,0.588,0.778}
\newcommand{\greg}[1]{{\textcolor{gregcolor}{[GS #1]}}}
\newcommand\ggs[1] {\greg{#1}}

\definecolor{seancolor}{rgb}{0.9,0.1,0.1}
\newcommand{\sean}[1]{{\textcolor{seancolor}{[SM #1]}}}
\newcommand\smo[1] {\sean{#1}}

%%%%%%%%% TITLE
\title{DeepLPF: Deep Local Parametric Filters for Image Enhancement}

\author{First Author\\
Institution1\\
Institution1 address\\
{\tt\small firstauthor@i1.org}
\and
Second Author\\
Institution2\\
First line of institution2 address\\
{\tt\small secondauthor@i2.org}
}

\author{Sean Moran$^{1}$, \,\,
Pierre Marza$^{\ast,1,2}$, \,\,
Steven McDonagh$^{1}$, \,\,
Sarah Parisot$^{1,3}$, \,\, 
Gregory Slabaugh$^{1}$ \\ 
 {\tt\small sean.j.moran@gmail.com, pierre.marza@gmail.com},\\{\tt\small \{steven.mcdonagh, sarah.parisot, gregory.slabaugh\}} 
{\tt\small @huawei.com},
\and
$^{1}$Huawei Noah's Ark Lab\\
\and
$^{2}$INSA Lyon
\and
$^{3}$Mila Montr\'{e}al
}

\maketitle
\thispagestyle{empty}

\blfootnote{
*Work done during an internship at Huawei Noah's Ark Lab.
}

\begin{abstract}

Digital artists often improve the aesthetic quality of digital photographs through manual retouching. Beyond global adjustments, professional image editing programs provide local adjustment tools operating on specific parts of an image. Options include parametric (graduated, radial filters) and unconstrained brush tools. These highly expressive tools enable a diverse set of local image enhancements. However, their use can be time consuming, and requires artistic capability. State-of-the-art automated image enhancement approaches typically focus on learning pixel-level or global enhancements. The former can be noisy and lack interpretability, while the latter can fail to capture fine-grained adjustments. In this paper, we introduce a novel approach to automatically enhance images using learned spatially local filters of three different types (Elliptical Filter, Graduated Filter, Polynomial Filter). We introduce a deep neural network, dubbed \emph{Deep Local Parametric Filters (DeepLPF)}, which regresses the parameters of these spatially localized filters that are then automatically applied to enhance the image.  DeepLPF provides a natural form of model regularization and enables interpretable, intuitive adjustments that lead to visually pleasing results.  We report on multiple benchmarks and show that DeepLPF produces state-of-the-art performance on two variants of the MIT-Adobe 5k~\cite{bychkovsky2011learning} dataset, often using a fraction of the parameters required for competing methods. 

\end{abstract}
\section{Introduction}
\label{sec:intro}

Digital photography has progressed dramatically in recent years due to sustained improvements in camera sensors and image signal processing pipelines.  Yet despite this progress, captured photographs may still lack quality due to varying factors including scene condition, poor illumination, or photographer skill. Human image retouchers often improve the aesthetic quality of digital photographs through manual adjustments. Professional-grade software (\eg Photoshop, Lightroom) allows application of a variety of modifications through both interactive and semi-automated tools.

\begin{figure}[t]
\includegraphics[scale=0.35]{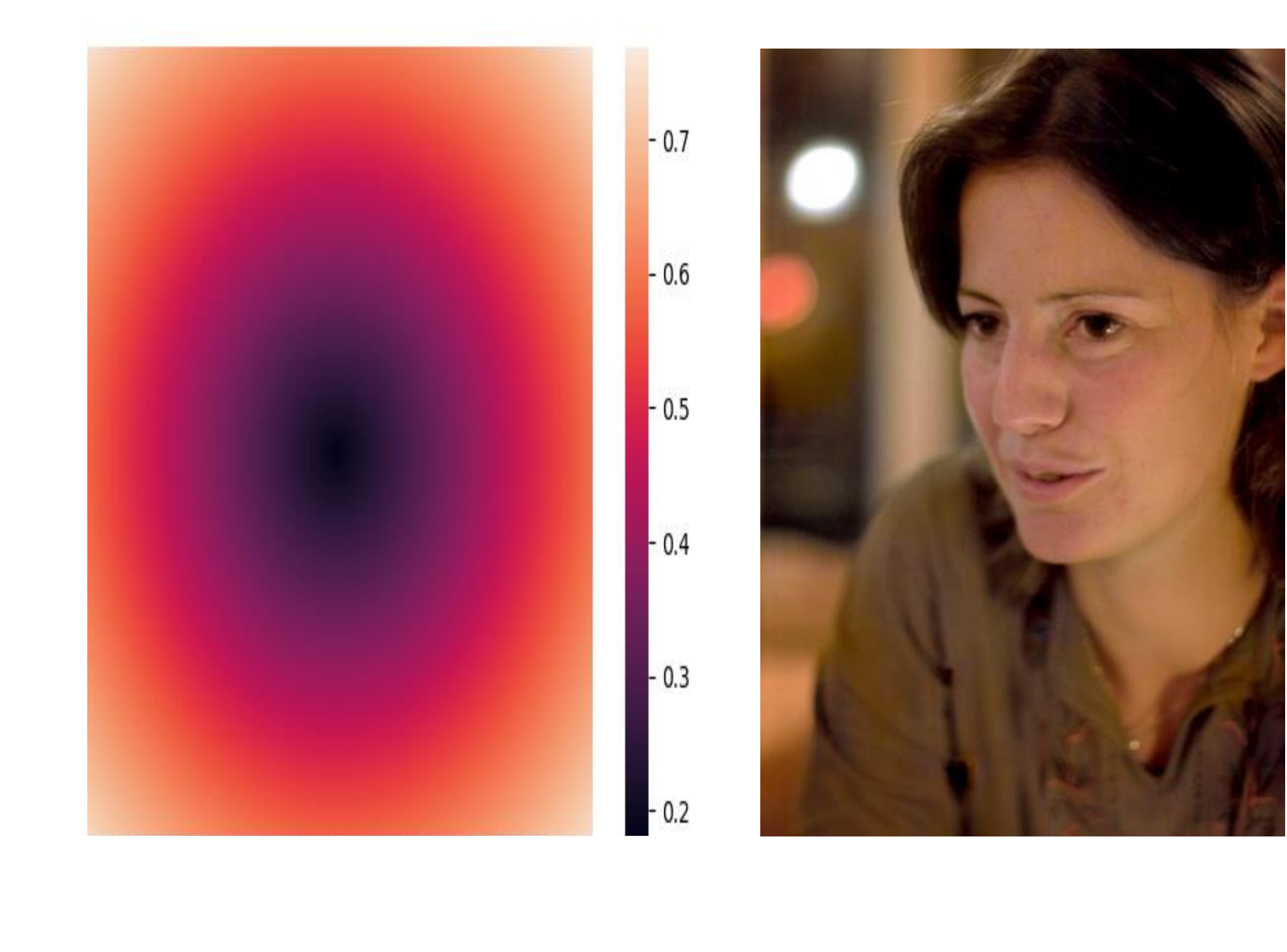}
\includegraphics[scale=0.28]{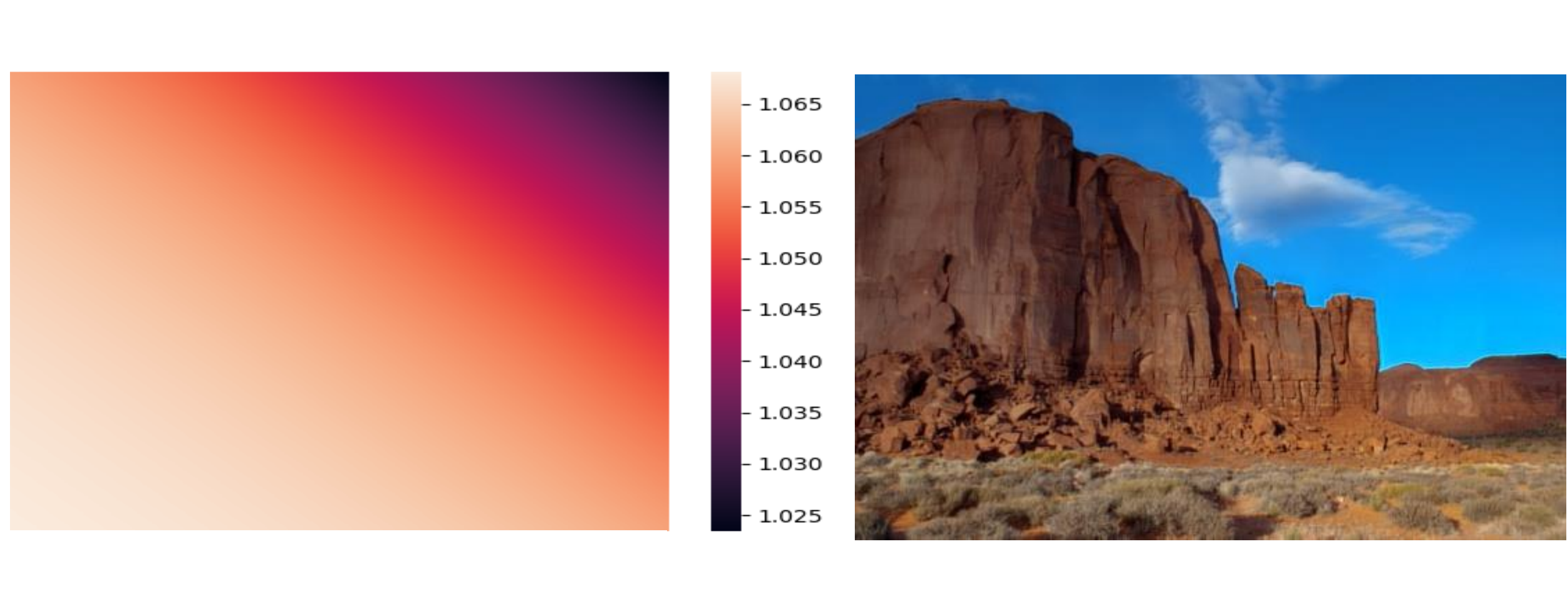}
\caption{DeepLPF for parametric local image enhancement.  \textbf{Left:} Examples of estimated filters. \textbf{Right:} The produced output images.
\textbf{Top:} Adjustment of the image red channel with a single Elliptical Filter. \textbf{Bottom:} Adjustment of the image red channel with a single Graduated Filter.}
\label{sec1:fig}
\end{figure}

In addition to elementary global adjustments such as contrast enhancement and brightening, advanced editing functionality is also available through local image adjustments, such as the examples shown in Fig~\ref{sec1:fig}. However, manual enhancement remains challenging for non-experts who may lack appropriate skills, time, or aesthetic judgment to improve their images effectively. 

These observations motivate the development of fully automatic photo enhancement tools that can replace non-expert user work or provide an improved manual-editing starting point for professional artists. Photographers often retouch images using a combination of different \emph{local filters} that only affect limited spatial regions of the image. For example, a photographer might want to adjust the darkness of the sky using a graduated filter while increasing the brightness of a face using an appropriately sized elliptical filter, and retouching small fine image detail using a brush tool.  

Inspired by this manual workflow, we propose a novel approach that learns parametric filters for local image enhancement. We draw influence from digital photo editing software, but model and emulate local editing tools using a deep neural network. Given (input, enhanced) pairs as training examples, we reproduce local, mid-level adjustments through learned graduated and elliptical filters and a learned brush tool. By constraining our model to learn how to utilize tools that are similar to those found in a digital artists' toolbox, we provide a natural form of model regularization and enable interpretable, intuitive adjustments that lead to visually pleasing results. 
Extensive experiments on multiple public datasets show that we outperform state-of-the-art~\cite{chen2018deep,wang2019underexposed,chen2018learning} results with a fraction of the neural network weight capacity. 

Our \textbf{Contributions} can be summarised as follows:

\begin{itemize}

\item \textbf{Local Parametric Filters}: We propose a method for automatic estimation of parametric filters for local image enhancement. We instantiate our idea using Elliptical, Graduated, Polynomial filters. Our formulation provides intuitively interpretable and intrinsically regularised filters that ensure weight efficiency (capacity frugality) and mitigate overfitting.

\item \textbf{Multiple Filter Fusion Block}: We present a principled strategy for the fusion of multiple learned parametric image filters. Our novel plug-and-play neural block is capable of fusing multiple independent parameter filter outputs and provides a flexible layer that can be integrated with common network backbones for image quality enhancement.

\item \textbf{State-Of-The-Art Image Enhancement Quality}: DeepLPF provides state-of-the-art image quality enhancement on two challenging benchmarks.
\end{itemize}

\section{Related work}
\label{sec:related_work}

Digital photo enhancement has a rich history in both the image processing and computer vision communities. Early automated enhancement focused primarily on image contrast~\cite{pizer1990contrast,stark2000adaptive,yuan2012automatic}, while recent work has employed data-driven methods to learn image adjustments for improving contrast, colour, brightness and saturation~\cite{hwang2012context,ignatov2017dslr,gharbi2017deep,chen2018deep,hu2018exposure,park2018distort}. The related image enhancement work can be broadly divided into methods that operate globally or locally on an image, or propose models that operate over both scales.

\textbf{Global image enhancement:} Bychkovsky \etal~\cite{bychkovsky2011learning} collected the popular MIT-Adobe-5K dataset\footnote{\url{https://data.csail.mit.edu/graphics/fivek/}} that consists of 5,000 photographs and their retouching by five different artists. The authors propose a regression based approach to learn the artists' photographic adjustments from image pairs. To automate colour enhancement, \cite{yan2014learning} propose a learning-to-rank approach over ten popular global colour controls. In~\cite{chen2017fast} an FCN is used to learn approximations to various global image processing operators such as photographic style, non-local dehazing and pencil drawing. Photo post-processing is performed by a white-box framework in~\cite{hu2018exposure} where global retouching curves are predicted in RGB space. A Reinforcement Learning (RL) approach enables the image adjustment process in~\cite{hu2018exposure} and Deep RL is used to define an ordering for enhancement adjustments in~\cite{park2018distort}, enabling application of global image modifications (\emph{eg.} contrast, saturation). 
 
\textbf{Local image enhancement:} Aubry~\etal~\cite{Aubry14} propose fast local Laplacian filtering for enhancing image detail. Hwang \etal~\cite{Hwang12} propose a method for local image enhancement that searches for the best local match for an image in a training database of (input, enhanced) image pairs.  Semantic maps are constructed in~\cite{yan2016automatic} to achieve semantic-aware enhancement and learn local adjustments. Underexposed photo enhancement~\cite{wang2019underexposed} (DeepUPE) learns a scaling luminance map using an encoder-decoder setup however no global adjustment is performed. The learned mapping has high complexity and crucially depends on the regularization strategy employed. Chen~\etal~\cite{chen2017fast} propose a method for fast learning of image operators using a multi-scale context aggregation network. Enhancement via image-to-image translation using both cycle consistency~\cite{zhu2017unpaired} and unsupervised learning~\cite{liu2017unsupervised} have also been proposed in recent years.

\textbf{Global and local image enhancement:} Chen~\etal~\cite{chen2018deep} develop Deep Photo Enhancer (DPE), a deep model for enhancement based on two-way generative adversarial networks (GANs). As well as local pixel-level adjustments, DPE introduces a global feature extraction layer to capture global scene context. Ignatov et al.~\cite{ignatov2018wespe} design a weakly-supervised image-to-image GAN-based network, removing the need for pixel-wise alignment of image pairs. Chen et al.~\cite{chen2018learning} propose a low-light enhancement model that operates directly on raw sensor data and propose a fully-convolutional approach to learn short-exposure, long-exposure mappings using their low-light dataset.  Recent work makes use of multitask learning~\cite{kong2019multitask} for real-time image processing with various image operators. Bilateral guided joint upsampling enables an encoder/decoder architecture for local as well as global image processing. HDRNet~\cite{gharbi2017deep} learns global and local image adjustments by leveraging a two stream convolutional architecture. The local stream extracts local features that are used to predict the coefficients of local affine transforms, while the global stream extracts global features that permit an understanding of scene category, average intensity \etc. In~\cite{ignatov2017dslr} DSLR-quality photos are produced for mobile devices using residual networks to improve both colour and sharpness.

In contrast with prior work, we propose to frame the enhancement problem by learning parametric filters, operating locally on an image. Learnable, parametrized filters align well with the intuitive, well-understood human artistic tools often employed for enhancement and this naturally produces appealing results that possess a degree of familiarity to human observers. Additionally, the use of our filter parameterization strategy serves to both constrain model capacity and regularize the learning process, mitigating over-fitting and resulting in moderate model capacity cost.
\section{Deep Local Parametric Filters (DeepLPF)}
\label{sec3}

\begin{figure}[t]
\begin{center}
\includegraphics[width=0.9\linewidth]{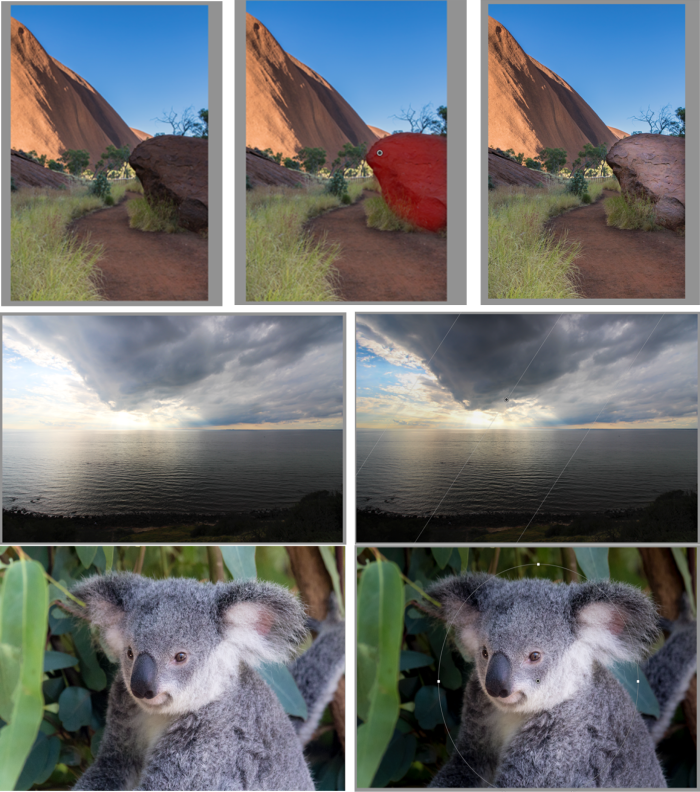}
\end{center}
\caption{Examples of local filter usage in  Lightroom for the brush tool (top row), graduated (middle row) and radial (bottom row) filters. Images are shown before (left) and after filter based enhancement by a human artist (right).}
\label{fig:lightroom}
\end{figure}

DeepLPF defines a novel approach for local image enhancement, introducing a deep fusion architecture capable of combining the output from learned, spatially local parametric image filters that are designed to emulate the combined application of analogous manual filters.
In this work, we instantiate three of the most popular local image filters (Elliptical, Graduated, Polynomial).
Figure~\ref{fig:lightroom} illustrates usage, and resulting effects of, comparable manual Lightroom filters. In Section~\ref{sec3:arch} we present our global DeepLPF architecture, designed to learn and apply sets of different parametric filters. Section~\ref{sec3:filters} then provides detail on the design of the three considered local parametric filters, and describes the parameter prediction block used to estimate filter parameters. Finally, Sections~\ref{sec3:fusing_filters} and~\ref{sec3:loss_fn} explain how multiple filters
are fused together and provide detail on our training loss function, respectively.

\begin{figure*}[t]
\begin{center}
\includegraphics[width=0.95\linewidth]{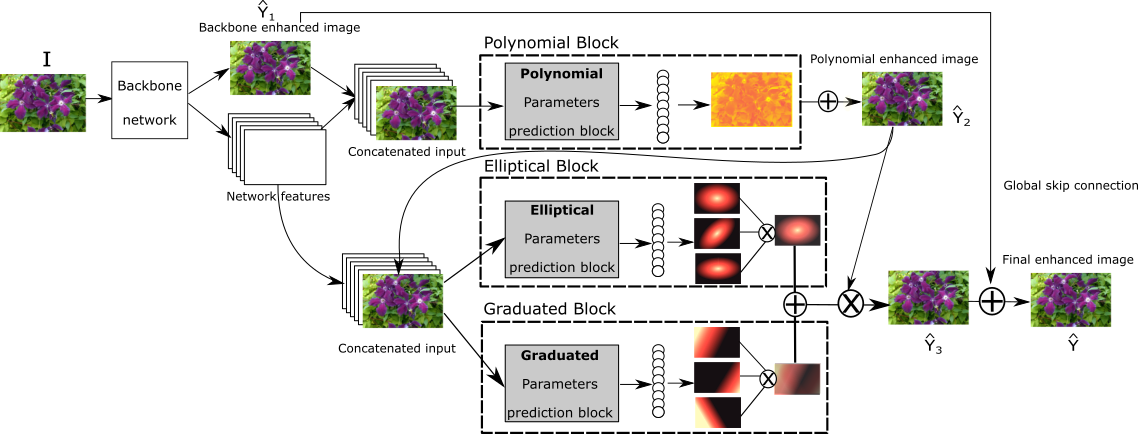}
\end{center}
\caption{Architecture diagram illustrating our approach to combine the different filter types (polynomial, elliptical, graduated) in a single end-to-end trainable neural network. The architecture combines a \emph{single stream path} for initial enhancement with the polynomial filter and a \emph{two stream path} for further refinement with the graduated and elliptical filters. }
\label{fig:deep_fusion_architecture}
\end{figure*}

\subsection{DeepLPF Architecture}
\label{sec3:arch}

The architecture of DeepLPF is shown in Figure~\ref{fig:deep_fusion_architecture}. 
Given a low quality RGB input image $I$ and its corresponding high quality enhanced target image $Y$, DeepLPF is trained to learn a transformation $f_{\theta}$ such that $\hat{Y}{=}f_{\theta}(I)$ is close to $Y$ as determined by an objective function based on image quality. Our model combines a single-stream network architecture for fine-grained enhancement, followed by a two-stream architecture for higher-level, local enhancement. We first employ a standard CNN backbone (\eg ResNet, UNet) to estimate a $H{\times}W{\times}C$ dimensional feature map. The first three channels of the feature map represent the image to be adjusted, while the remaining $C'{=}C{-}3$ channels  
represent additional features 
that feed into three filter parameter prediction blocks. The first \emph{single stream path} estimates the parameters of a polynomial filter that is subsequently applied to the pixels of the backbone enhanced image $\hat{Y}_{1}$
input image. This block emulates a brush tool which adjusts images at the pixel level with a smoothness constraint imposed by the brush's shape. The image enhanced by the polynomial filter, $\hat{Y}_{2}$, is concatenated with the $C'$ backbone features, and serves as input to the \emph{two stream path}, which learns and applies more constrained, local enhancements in the form of elliptical and graduated filters. The adjustment maps of the elliptical and graduated filters are estimated using two parallel regression blocks. 
Elliptical and graduated maps are fused using simple addition, although more involved schemes could be employed \eg weighted combination. This fusion step results in a scaling map $\hat{S}$ that is element-wise multiplied to $\hat{Y}_{2}$ to give image $\hat{Y}_{3}$, effectively applying the elliptical and graduated adjustments to the image after polynomial enhancement. The image $\hat{Y}_{1}$, enhanced by the backbone network, is finally added through a long residual connection to $\hat{Y}_{3}$ producing the final output image $\hat{Y}$.

\subsection{Local Parametric Filters}
\label{sec3:filters}

We provide detail on three examples of local parametric filter: the Graduated Filter (\ref{sec3:filters:graduated}), Elliptical Filter (\ref{sec3:filters:elliptical}) and Polynomial Filter (\ref{sec3:filters:polynomail}). Filters permit different types of local image adjustment, governed by their parametric form, and parameter values specify the exact image effect in each case. Filter parameter values are image specific and are predicted using supervised CNN regression (\ref{sec:parameter_prediction}). 

\subsubsection{Filter Parameter Prediction Network}
\label{sec:parameter_prediction}

Our parameter prediction block is a lightweight CNN 
that accepts a feature set from a backbone network and regresses filter parameters individually. The network block alternates between a series of convolutional and max pooling layers that gradually downsample the feature map resolution. Following these layers there is a global average pooling layer and a fully connected layer which is responsible for predicting the filter parameters. The global average pooling layer ensures that the network is agnostic to the resolution of the input feature set. Activation functions are Leaky ReLUs and dropout ($50\%$, both train and test) is applied to the fully connected layer. Further architectural details are provided in our supplementary material.

For the three example filters considered; the only difference between respective network regressors is the number of output nodes in the final fully connected layer. This corresponds to the number of parameters that define the respective filter (Table~\ref{tab:params}). In comparison to contemporary local pixel level enhancement methods~\cite{chen2018deep}, the number of parameters required is greatly reduced.  
Our network output size can be altered to estimate the parameters of multiple instances of the same filter type. Our methodology for predicting the parameters of an image transformation is aligned with previous work~\cite{gharbi2017deep,Chen16,Shih13} that shows that learning a parameterised transformation is often more effective and simpler than directly predicting an enhanced image directly. Importantly, the regression network is agnostic to the specific implementation of backbone model allowing image filters to be used to enhance the output of any image translation network.

\begin{table}[t]
\caption{Parameters used in localized filters}
\centering
\begin{adjustbox}{max width=0.495\textwidth}
\begin{tabular}{lll}
\multicolumn{3}{c}{} \\
\textbf{Filter} &  \textbf{\# Parameters} &  \textbf{Parameters}  \\
\hline
Graduated       & $G{=}8$  & $s^R_g, s^G_g, s^B_g, m, c, o_1, o_2, g_{inv}$ \\ 
Elliptical      & $E{=}8$  & $s^R_e, s^G_e, s^B_e, h, k, \theta, a, b$      \\
Cubic-10 & $P{=}30$ & $\{A \cdots J\}$ per colour channel             \\
Cubic-20 & $P{=}60$ & $\{A \cdots T\}$ per colour channel             \\
\end{tabular}
\end{adjustbox}
\label{tab:params}
\end{table}

\subsubsection{Graduated Filter}
\label{sec3:filters:graduated}
\begin{figure}
\centering
\subcaptionbox{\label{fig:graduated_filter}}{ \includegraphics[width=0.48\linewidth]{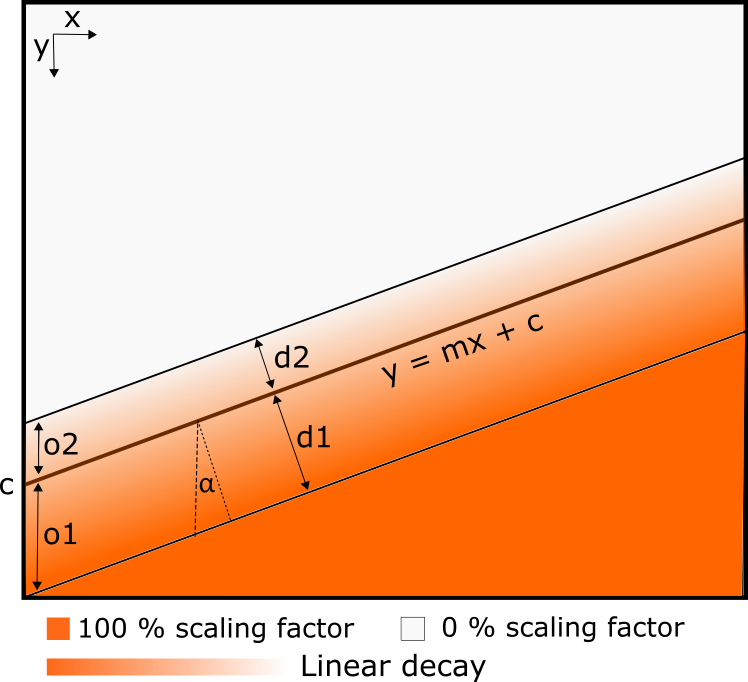}
}%
\hfill
\subcaptionbox{\label{fig:elliptical_filter}}{
\includegraphics[width=0.48\linewidth]{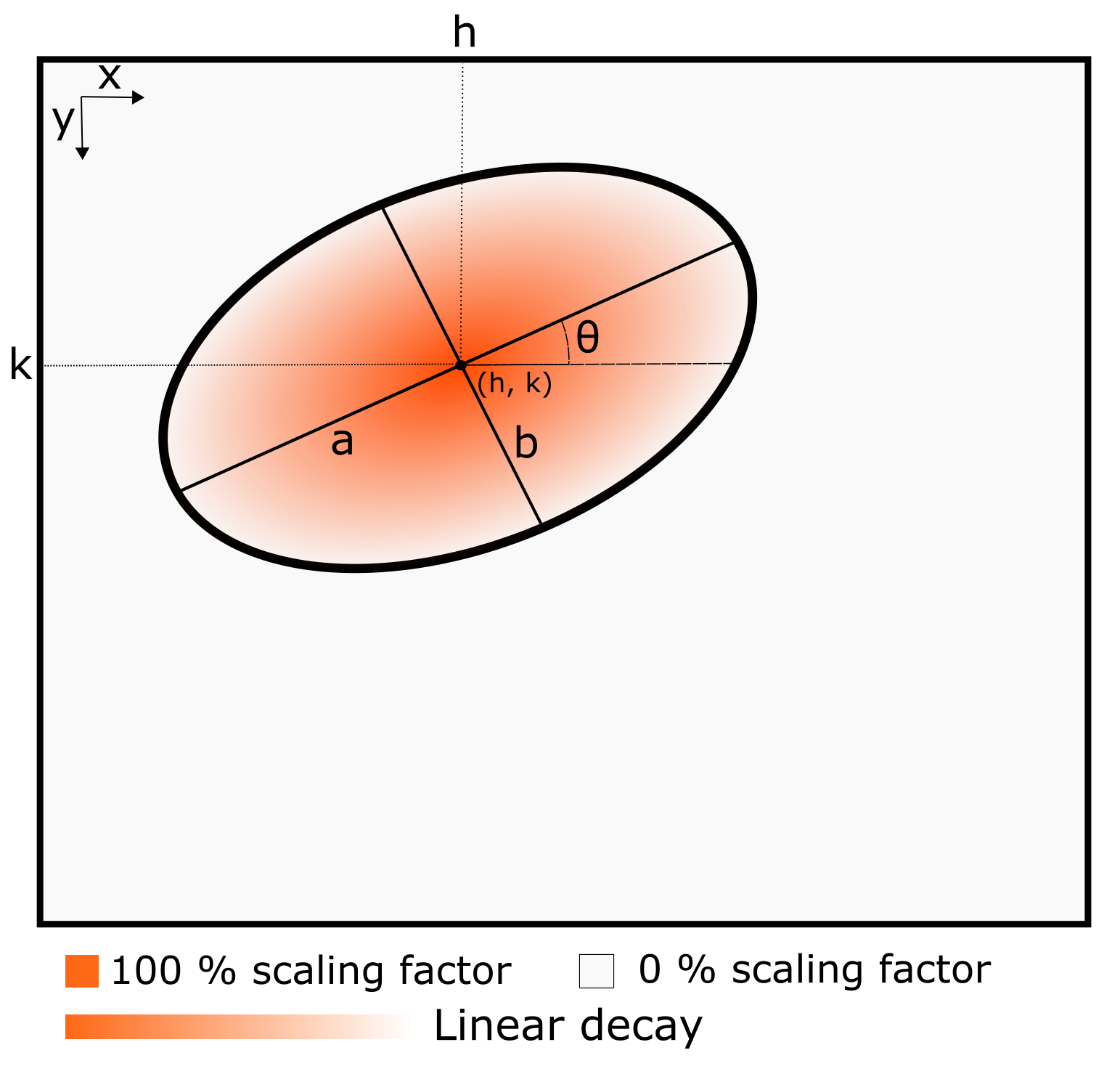}}%
\caption{Parametrisation and heatmap of the graduated (a) and elliptical (b) filters.  See main text for further detail.}
\label{fig:graduated_elliptical_filter}
\end{figure}

Graduated filters are commonly used in photo editing software to adjust images with high contrast planar regions such as an overexposed sky. 
Our graduated image filter, illustrated in Figure~\ref{fig:graduated_filter}, is parametrised by three parallel lines. The central line defines the filter location and orientation, taking the form $y{=}mx+c$ with slope $m$ and intercept $c$, providing a linear parameterisation in standard fashion. Offsets 
$o_1$ and 
$o_2$ provide two additional parameters such that each (channel-wise) adjustment map is composed of four distinct regions that form a heatmap $s(x,y)$. In the the 100\% area all pixels are multiplied by scaling parameter $s_g$. 
Then, in the 100-50\% area, the applied scaling factor linearly decreases from $s_g$ to $\frac{s_g}{2}$. Inside the 50-0\% area, the scaling value is further decreased linearly until reaching the 0\% area where pixels are not adjusted.  Mathematically, the graduated filter is described in Equations \ref{eq:graduated_filter1}-\ref{eq:graduated_filter3}:

\begin{equation}
    a(x,y) = s_g \min \left\{ \frac{1}{2} \left(1 + \frac{\ell(x,y)}{d_2} \right), 1\right\}
\label{eq:graduated_filter1}
\end{equation}
 
\begin{equation}
    b(x,y) = s_g \max \left\{ \frac{1}{2} \left(1 + \frac{\ell(x,y)}{d_1} \right), 0\right\}
\label{eq:graduated_filter2}
\end{equation}
 
\begin{equation}
    s(x,y) = \left\{ \begin{array}{ll} \hat{g}_{inv}a(x,y) +(1-\hat{g}_{inv})b(x,y) & \hspace{-0.5em}, l(x,y) \ge 0 \\
    (1-\hat{g}_{inv})a(x,y) +\hat{g}_{inv}b(x,y) & \hspace{-0.5em}, l(x,y) < 0 \end{array} \right.
\label{eq:graduated_filter3}
\end{equation}
where $\ell(x,y) = y - (mx+c)$ is a function of the location of a point $(x,y)$ relative to the central line, $d_1 = o_1 \cos \alpha$, $d_2 = o_2 \cos \alpha$, $\alpha = \tan^{-1}(m)$, and $\hat{g}_{inv}$ is a binary indicator variable. Parameter $\hat{g}_{inv}$ permits \emph{inversion} with respect to the top and bottom lines. Inversion determines the location of the 100\% scaling area relative to the central line. To enable learnable inversion, we predict a binary indicator parameter $\hat{g}_{inv}=\frac{1}{2}(sgn(g_{inv})+1)$, where $sgn$ denotes the sign function, $g_{inv}$ is the real-valued predicted parameter and $\hat{g}_{inv}$ is the binarised $\hat{g}_{inv}\in\{0,1\}$ version. The gradient of this sign function is everywhere zero and undefined at zero, therefore we use the straight-through estimator~\cite{Bengio13} on the backward pass to learn this binary variable. 

\subsubsection{Elliptical Filter}
\label{sec3:filters:elliptical}

A further filter we utilise defines an ellipse parametrised by the center $(h, k)$, semi-major axis ($a$), semi-minor axis ($b$) and rotation angle ($\theta$). The learned scaling factor $s_e$ is maximal at the center of the ellipse (100\% point) and decreases linearly until reaching the boundary. Pixels are not adjusted outside the ellipse ($0\%$ area).

Mathematically, a channel-wise heatmap, defined by the elliptical filter, is denoted as:

\begin{eqnarray}
s(x,y) & = & s_e \min \left(0, 1 - \frac{\left[(x-h)\cos \theta + (y-k)\sin \theta\right]^2}{a^2} \right. \nonumber \\
& &  + \left. \frac{\left[(x-h)\sin \theta - (y-k)\cos \theta\right]^2}{b^2} \right).
\end{eqnarray}

An example elliptical filter parametrisation and heatmap is illustrated in Fig.~\ref{fig:elliptical_filter}. Elliptical filters are often used to enhance, accentuate objects or specific regions of interest within photographs \eg human faces.

\subsubsection{Polynomial Filter}
\label{sec3:filters:polynomail}

Our third considered filter type constitutes a polynomial filter capable of providing fine-grained, regularised adjustments over the entire image. The polynomial filter emulates a brush tool, providing a broad class of geometric shapes while incorporating spatial smoothness.  We consider order-$p$ polynomial filters of the forms $i{\cdot}(x + y + \gamma)^p$ and $(x + y + i + \gamma)^p$, where $i$ is the image channel intensity at pixel location $(x,y)$, and $\gamma$ is an independent scalar.  
We empirically find a cubic polynomial ($p=3$) to offer both expressive image adjustments yet only a limited set of parameters. We explore two variants of the cubic filter, \emph{cubic-10} and \emph{cubic-20}.

Our smaller cubic filter (\emph{cubic-10}) constitutes a set of $10$ parameters to predict; $\{A,B,C,\ldots,J\}$, defining a cubic function $f$ that maps intensity $i$ to a new adjusted intensity $i'$:

\begin{equation}
\begin{aligned}
i'(x, y)  
 &= f(x, y, i) \\
   &= i*(A x^3 + B x^2 y + C x^2 + D  +  \\
   &+ D x y^2 + E x y + F x  + G y^3  + H y^2   \\
   &  + I y  + J) \\
\end{aligned}
\label{eq:cubic10}
\end{equation}

Considering twice as many learnable parameters, the \emph{cubic-20} filter explores higher order intensity terms, and consists of a set of $20$ parameters $\{A,B,C,\ldots,T\}:$ 
\begin{equation}
\begin{aligned}
i'(x, y) &= f(x, y, i) \\
   &= Ax^3 + Bx^2y + Cx^2i + Dx^2  + Exy^2 \\
   &+ Fxyi + Gxy   + Hxi^2 + Ixi   + Jx    \\
   &+ Ky^3 + Ly^2i + My^2  + Nyi^2 + Oyi   \\
   &+ Py   + Qi^3  + Ri^2  + Si    + T \\
\end{aligned}
\label{eq:cubic20}
\end{equation}

Our cubic filters consider \emph{both} spatial and intensity information while constraining the complexity of the learnt mapping to a regularized form. This permits the learning of precise, pixel level enhancement while ensuring that transformations are locally smooth. We estimate an independent cubic function for each colour channel yielding a total of $30$ parameters for the \emph{cubic-10} and $60$ parameters for the \emph{cubic-20} filter respectively.

\begin{figure*}[t]
\begin{center}

\begin{tabular}{c@{}c@{}c@{}}
    \scalebox{1}{Elliptical (Green Channel)} &
      \scalebox{1}{Graduated (Blue Channel)} & 
      \scalebox{1}{Cubic (Red Channel)} \\
     \includegraphics[scale=0.23]{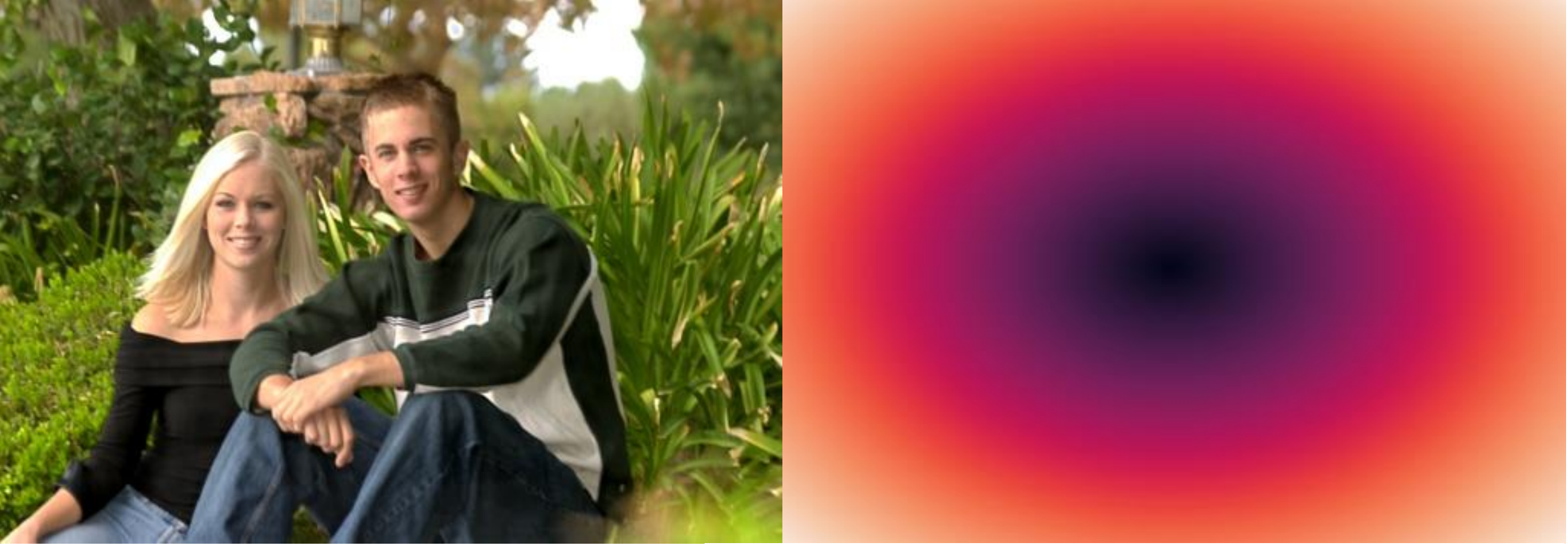}&
    \includegraphics[scale=0.23]{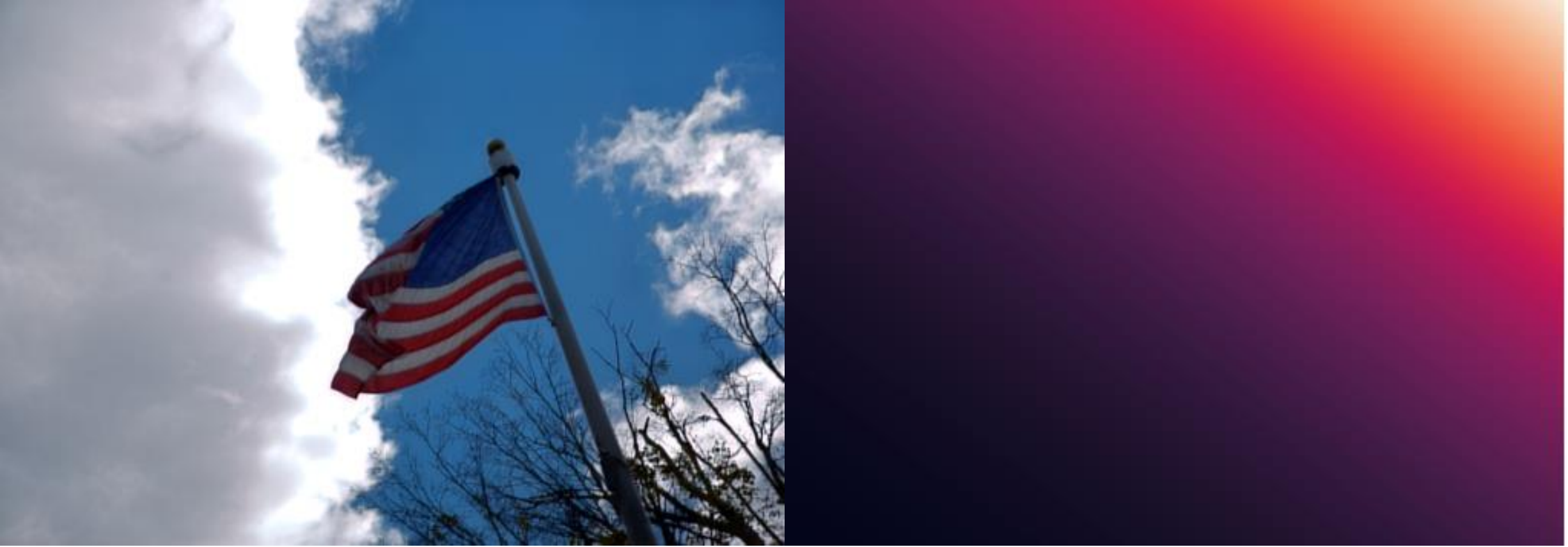} &
    \includegraphics[scale=0.23]{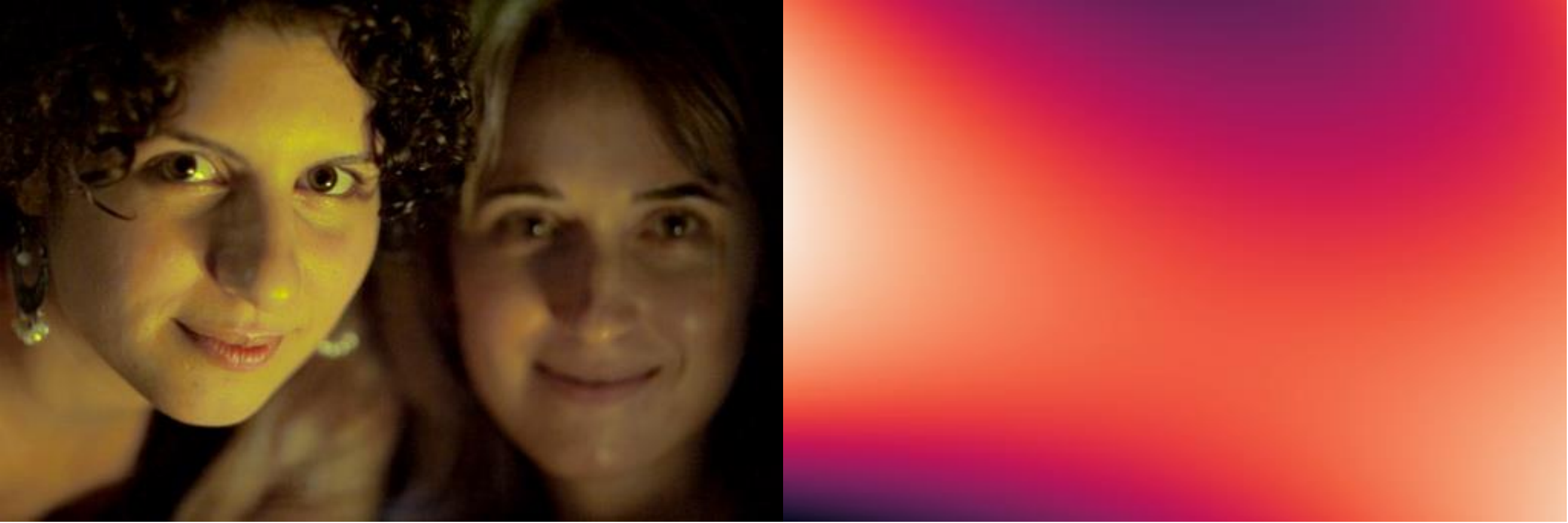} \\
\end{tabular}
\end{center}
\caption{Model output images and examples of Elliptical, Graduated and Polynomial (cubic-10) image filters learnt by one instance of DeepLPF. Lighter filter heat map colours correspond to larger image adjustment values. Cubic-10 filter is shown without the intensity ($i$) term. Best viewed in colour.}
\label{fig:filter_comparison1}
\end{figure*}

\subsection{Fusing Multiple Filters of the Same Type}
\label{sec3:fusing_filters}

Our graduated and elliptical prediction blocks can each output parameter values for $n$ instances of their respective filter type. Further exploration towards choosing $n$ appropriately is detailed in Section~\ref{sec:experiments:results}. In the case $n > 1$, multiple instances are combined into an adjustment map through element-wise multiplication,
\begin{equation}
\begin{aligned}
s_g(x,y) = \prod_{i=1}^{n} s_{gi}(x,y), \quad
s_e(x,y) = \prod_{i=1}^{n} s_{ei}(x,y)
\end{aligned}
\end{equation}
where $s_{gi}(x,y)$ and $s_{ei}(x,y)$ are the adjustment map of the $i^{th}$ instance of the corresponding graduated and elliptical filters respectively. On the contrary, since a single per-channel cubic filter allows intrinsically high flexibility in terms of expressive power, we opt not to fuse multiple cubic filters together. 

\subsection{DeepLPF Loss Function}
\label{sec3:loss_fn}

The DeepLPF training loss leverages the CIELab colour space to compute the $L_{1}$ loss on the Lab channels and the MS-SSIM loss on the L channel (Equation~\ref{eq:dpf_loss}). By splitting chrominance and luminance information into separate loss terms, our model is able to account for separate focus on both local (MS-SSIM) and global ($L_{1}$) image enhancements during training~\cite{schwartz19,zhao2015functions}. Given a set of $N$ image pairs $\{(Y_{i},\hat{Y}_{i})\}^{N}_{i=1}$, where $Y_{i}$ is the reference image and $\hat{Y}_{i}$ is the predicted image, we define the DeepLPF training loss function as:
\vspace{-0.15in}

\begin{align}
\mathcal{L} &=  \sum^{N}_{i=1} \{ \omega_{\text{lab}}||Lab(\hat{Y}_{i})-Lab(Y_{i})||_{1}        \label{eq:dpf_loss}\\
       & + \omega_{\text{ms{-}ssim}}\text{MS-SSIM}(L(\hat{Y}_{i}),L(Y_{i})) \} \nonumber 
        \label{eq:dpf_loss}
\end{align}

\noindent where $Lab(\cdot)$ is a function that returns the CIELab Lab channels corresponding to the RGB channels of the input image and $L(\cdot)$ returns the L channel of the image in CIELab colour space. MS-SSIM is the multi-scale structural similarity~\cite{wang03}, and $\omega_{\text{lab}}, \omega_{\text{ms{-}ssim}}$ are hyperparameters weighting the relative influence of the terms in the loss function. 

%\section{Experimental Evaluation}
\section{Experiments}
\label{sec:experiments}

\subsection{Experimental Setup}
\label{sec:experiments:setup}

\textbf{Datasets:} We evaluate DeepLPF on three challenging benchmarks, derived from two public datasets. Firstly \textbf{(i) MIT-Adobe-5K-DPE~\cite{chen2018deep}}: $5000$ images captured using various DSLR cameras. Each captured image is subsequently (independently) retouched by five human artists. In order to make our results reproducible and directly comparable to the state-of-the-art, we consider only the (supervised) subset used by DeepPhotoEnhancer (DPE)~\cite{chen2018deep} and additionally follow their dataset pre-processing procedure. The image retouching of Artist C is used to define image enhancement ground truth. The data subset is split into $2250$, $500$ training and testing image pairs respectively. We randomly sample $500$ images from the training set, providing an additional validation set for hyperparameter optimisation. The images are resized to have a long-edge of $500$ pixels. \textbf{(ii) MIT-Adobe-5K-UPE~\cite{wang2019underexposed}:} our second benchmark consists of the same image content as \textbf{MIT-Adobe-5K-DPE} however image pre-processing here differs and instead follows the protocol of DeepUPE~\cite{wang2019underexposed}. We therefore refrain from image resizing and dataset samples vary in pixel resolution; $6\textnormal{-}25M$. We additionally follow the train / test split provided by~\cite{wang2019underexposed} and our second benchmark therefore consists of $4500$ training image pairs, from which we randomly sample $500$ to form a validation set. Testing images ($500$) are identical to those samples selected by DeepUPE and ground truth again consists of the Artist C manually retouched images. \textbf{(iii) See-in-the-dark (SID)}~\cite{chen2018learning}: the dataset consists of $5094$ image pairs, captured by a Fuji camera. For each pair the input is a short-exposure image in raw format and the ground truth is a long-exposure RGB image. Images are $24M$ pixel in size and content consists of both indoor and outdoor environments capturing diverse scenes and common objects of varying size.

\textbf{Evaluation Metrics:}
We evaluate quantitatively using PSNR, SSIM and the perceptual LPIPS metric~\cite{zhang18}. 

\textbf{Implementation details:} Our experiments all employ a U-Net backbone~\cite{Ronneberger15}. The base U-Net architecture, used for MIT-Adobe-5K-DPE experimentation, is detailed in the supplementary material. The U-Net architectures used for the other benchmarks are similar yet have a reduced number of convolutional filters (MIT-Adobe-5K-UPE) or include pixel shuffling functionality~\cite{Ledig17} to account for RAW input (SID). All experiments use the Adam Optimizer with a learning rate of $10^{-4}$. Our architecture makes use of three graduated (elliptical) filters per channel and we search for loss function (Eq.~\ref{eq:dpf_loss}) hyperparameters empirically resulting in: $\omega_{lab}{=}\{1, 1, 1\}$ and $\omega_{ms{-}ssim}{=} \{10^{-3}, 10, 10^{-3}\}$ for all MIT-Adobe-5K-DPE, MIT-Adobe-5K-UPE and SID experiments, respectively.

\begin{table}[t]
\centering
\caption{Model filter type ablation study (upper) and comparisons with state-of-the art methods (lower) using the \textbf{MIT-Adobe-5K-DPE} benchmark. PSNR and SSIM results, reported by competing works, are replicated from~\cite{chen2018deep}.} 
\begin{adjustbox}{max width=0.495\textwidth}
\begin{tabular}{ p{50mm} | c c c c }
\multicolumn{4}{c}{} \\
\hline
\textbf{Architecture} & \textbf{PSNR}$\uparrow$ & \textbf{SSIM}$\uparrow$ & \textbf{LPIPS}$\downarrow$ & \textbf{\# Weights} \\
\hline
U-Net                                       & ${21.57}$ & ${0.843}$ & 0.601  & 1.3 M \\  %fixed 12th feb
U-Net{+}Elliptical                          & ${22.56}$ & ${0.879}$ & ${-}$  & 1.5 M \\  %fixed 12th feb
U-Net{+}Graduated                                  & ${22.64}$ & ${0.888}$ & ${-}$  & 1.5 M \\  %fixed 12th feb
U-Net{+}Elliptical{+}Graduated             & ${22.88}$ & ${0.886}$ & ${-}$ &1.6 M   \\
U-Net{+}Cubic-10                            & ${22.69}$& ${0.871}$ & ${-}$  & 1.5 M  \\ 
U-Net{+}Cubic-20                            & ${23.44}$& ${0.886}$ & ${-}$  &  1.5 M \\ 
\small{U-Net{+}Cubic-20{+}Elliptical{+}Graduated} & $\textbf{23.93}$   & \textbf{0.903} & \textbf{0.582}  & \textbf{1.8 M} \\  
\hline
\hline
DPED~\cite{ignatov2017dslr}                        & ${21.76}$ & ${0.871}$ & ${-}$  &  ${-}$ \\ 
8RESBLK~\cite{zhu2017unpaired,liu2017unsupervised} & ${23.42}$ & ${0.875}$ & ${-}$  &  ${-}$ \\ 
FCN~\cite{chen2017fast}                            & ${20.66}$ & ${0.849}$ & ${-}$  &  ${-}$ \\ 
CRN~\cite{chen2017photographic}                    & ${22.38}$ & ${0.877}$ & ${-}$  &  ${-}$ \\  
%U-Net (DPE)~\cite{chen2018deep}                   & ${22.13}$ & ${0.879}$ & ${-}$  &  ${-}$ \\ 
DPE~\cite{chen2018deep}                            & ${23.80}$ & ${0.900}$ & 0.587  & 3.3 M  \\ 
\hline
\end{tabular}
\end{adjustbox}
\label{tab:results1}
\end{table}

\subsection{Experimental Results}
\label{sec:experiments:results}

\noindent \textbf{Ablation study:} 
We firstly conduct experiments to understand the different contributions and credit assignment for our method filter components, using the MIT-Adobe-5K-DPE data. Table~\ref{tab:results1} (upper) shows ablative results for our considered image filter components. For each configuration, we report PSNR metrics and verify experimentally the importance of integrating each component and their superior combined performance.

Individual filters are shown to bring boosts in performance \cf the U-Net backbone alone. The polynomial (cubic) filter has a greater impact than elliptical and graduated filters. We hypothesise that this can be attributed to the spatially large yet non-linear effects enabled by this filter type. Combining all filter blocks incorporates the benefits of each and demonstrates strongest performance. We highlight that only ${\sim}\frac{1}{4}$ of model parameters (\ie around $452k$) in the full architecture are attributed to filter blocks. The remaining capacity is dedicated to the U-Net backbone, illustrating that the majority of performance gain is a result of designing models capable of frugally emulating manual image enhancement tools.

\begin{figure}[t]
\centering
\subcaptionbox{\label{fig:ablation_filters:psnr}}{ \includegraphics[width=0.48\linewidth]{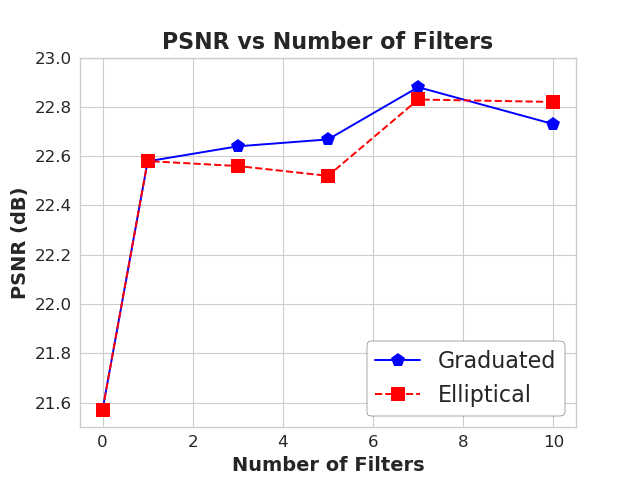}
}%
\hfill
\subcaptionbox{\label{fig:ablation_filters:ssim}}{\includegraphics[width=0.48\linewidth]{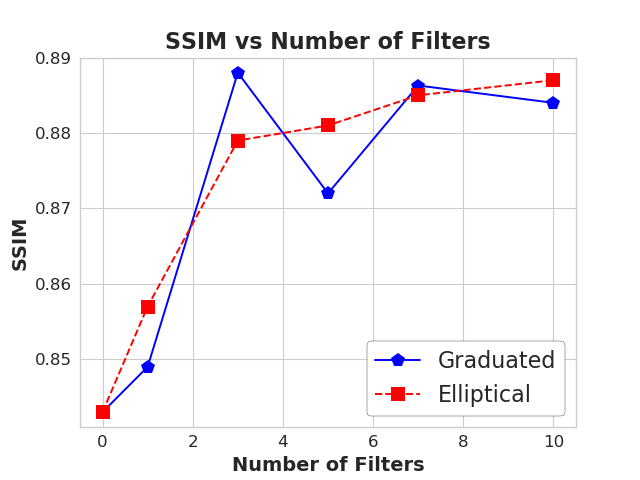}}%
\caption{Experimental study evaluating the effect of the number of graduated and elliptical filters used (\textbf{MIT-Adobe-5K-DPE dataset}).}
\label{fig:ablation_filters}
\end{figure}

We further investigate the influence of graduated and elliptical filter \emph{quantity} on model performance in Figure~\ref{fig:ablation_filters}. We find a general upward trend in PSNR, SSIM metrics as the number of filters per channel are increased. This upward trend can be attributed to the increased modelling capacity brought about by the additional filters. We select 3 filters per channel in our experiments which provides a tradeoff between image quality and parameter count.

\begin{figure*}[t]
\begin{center}
\begin{tabular}{c@{}c@{}c@{}c@{}}
    \scalebox{0.85}{Input} &
      \scalebox{0.85}{DPE~\cite{chen2018deep}} & 
      \scalebox{0.85}{\textbf{DeepLPF} } &
      \scalebox{0.85}{Ground Truth} \\
    \includegraphics[width=0.2\linewidth]{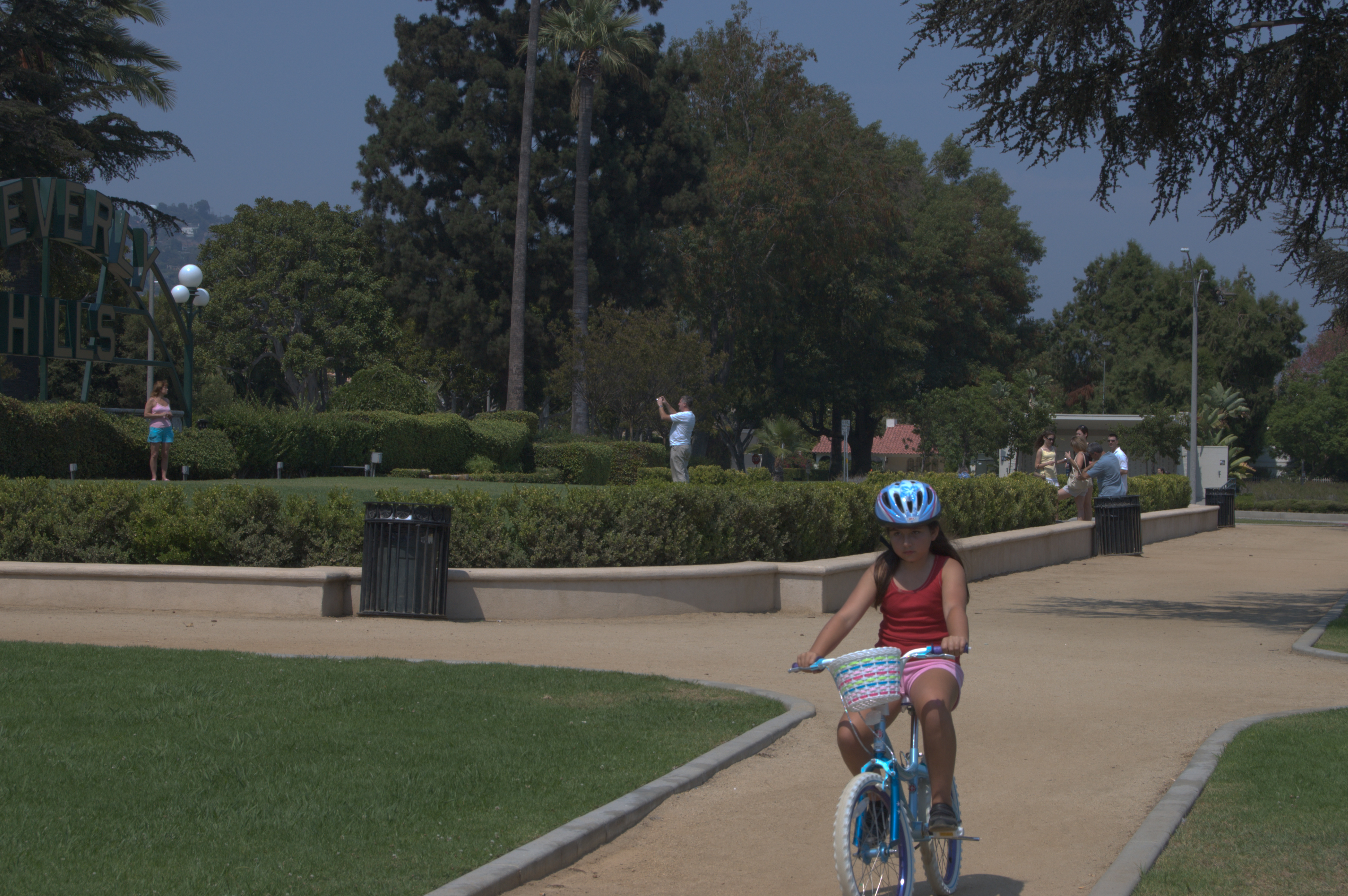} &
    \includegraphics[width=0.2\linewidth]{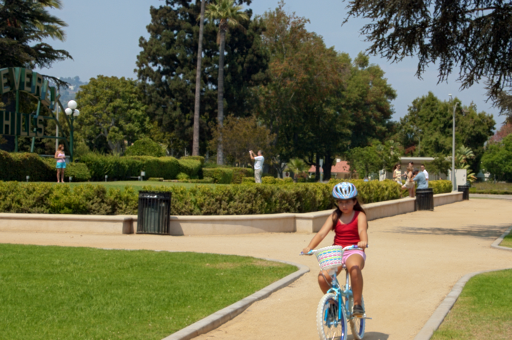} &
    \includegraphics[width=0.2\linewidth]{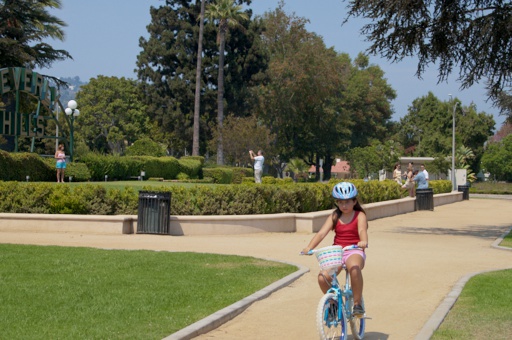} &
    \includegraphics[width=0.2\linewidth]{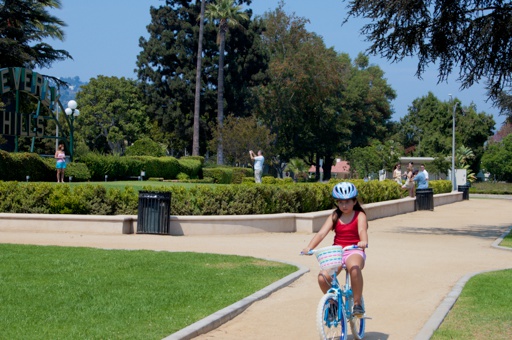}\\
\end{tabular}
\begin{tabular}{c@{}c@{}c@{}c@{}}
    \scalebox{0.85}{Input} &
      \scalebox{0.85}{DeepUPE~\cite{chen2018deep}} & 
      \scalebox{0.85}{\textbf{DeepLPF} } &
      \scalebox{0.85}{Ground Truth} \\
     \includegraphics[width=0.2\linewidth]{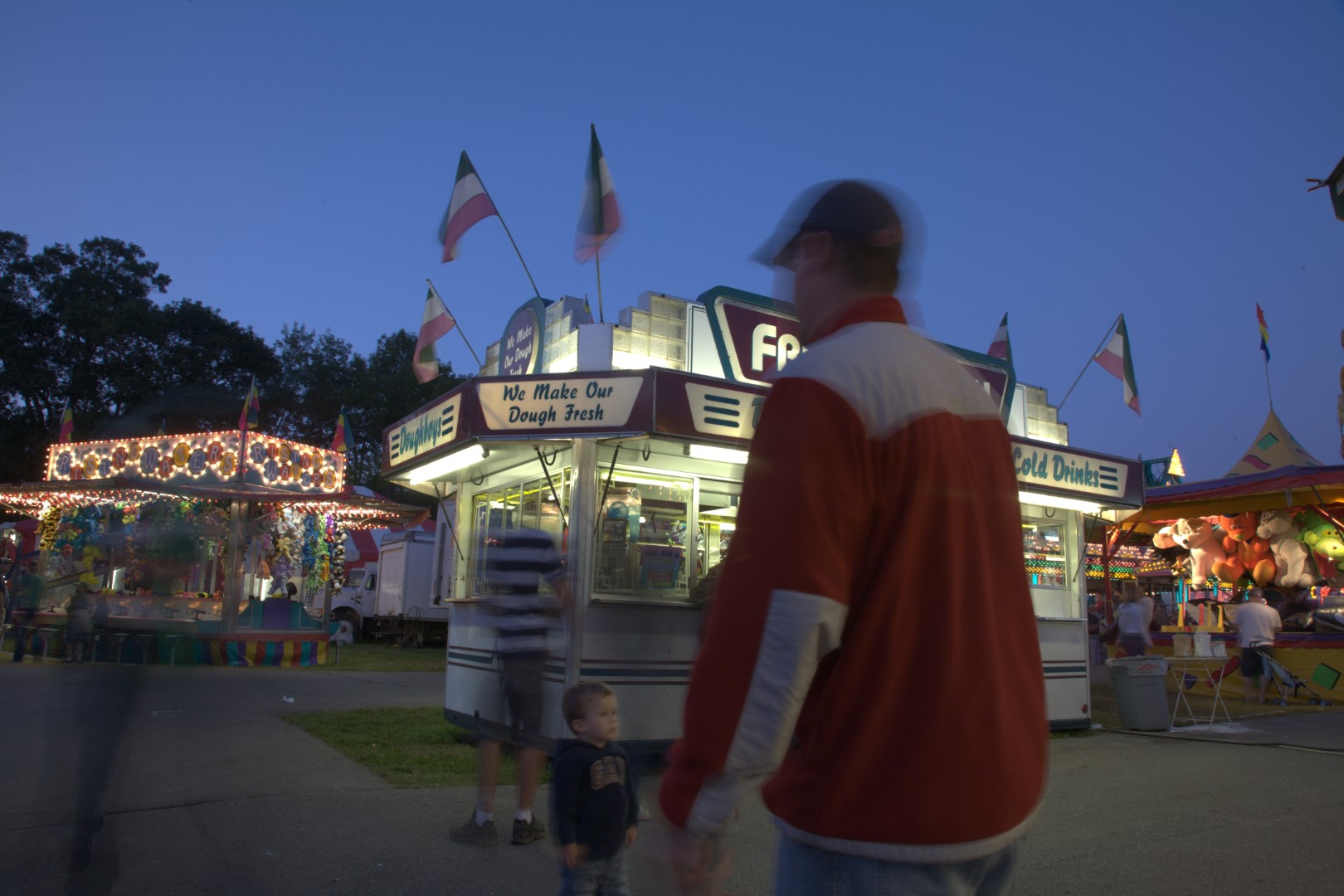}&
\includegraphics[width=0.2\linewidth]{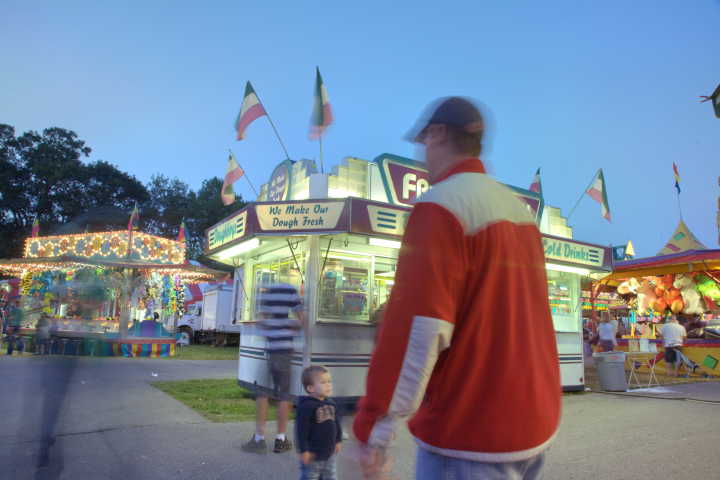}&
\includegraphics[width=0.2\linewidth]{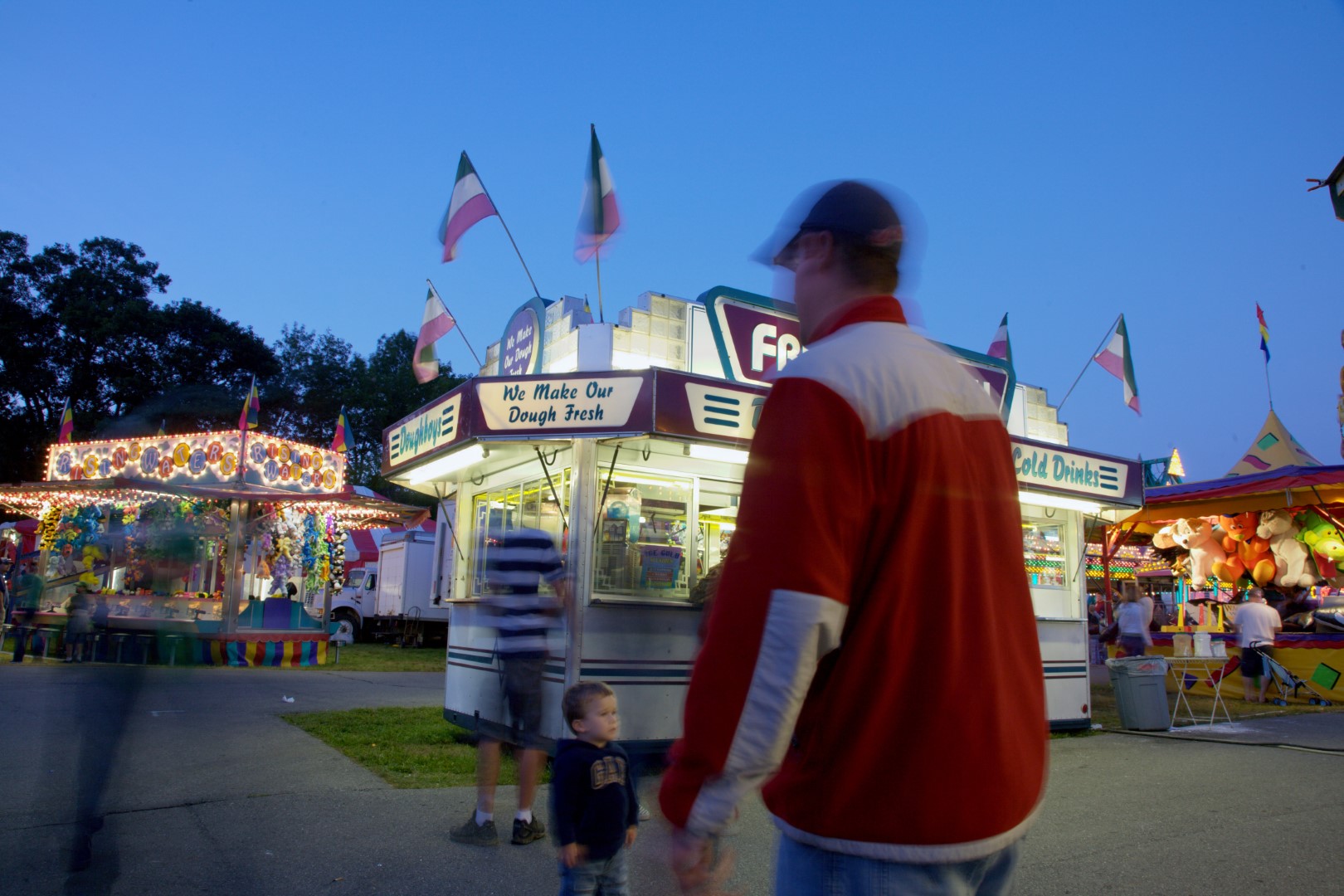}&
\includegraphics[width=0.2\linewidth]{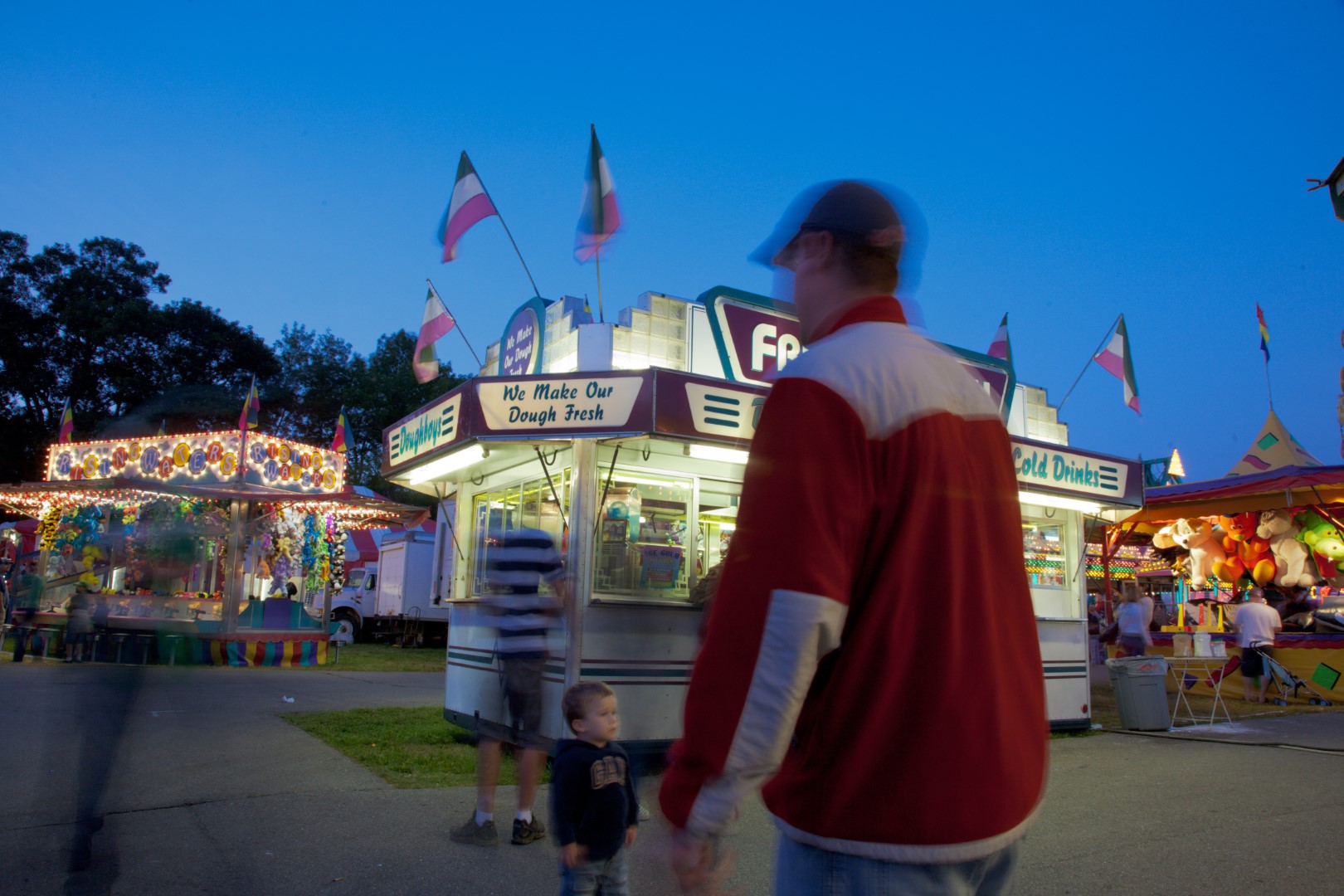} \\
\end{tabular}
\begin{tabular}{c@{}c@{}c@{}c@{}}
\scalebox{0.85}{Input} & 
\scalebox{0.85}{SID (U-Net)~\cite{chen2018learning}} & 
\scalebox{0.85}{\textbf{DeepLPF} } &
\scalebox{0.85}{Ground Truth} \\
\includegraphics[width=0.2\linewidth]{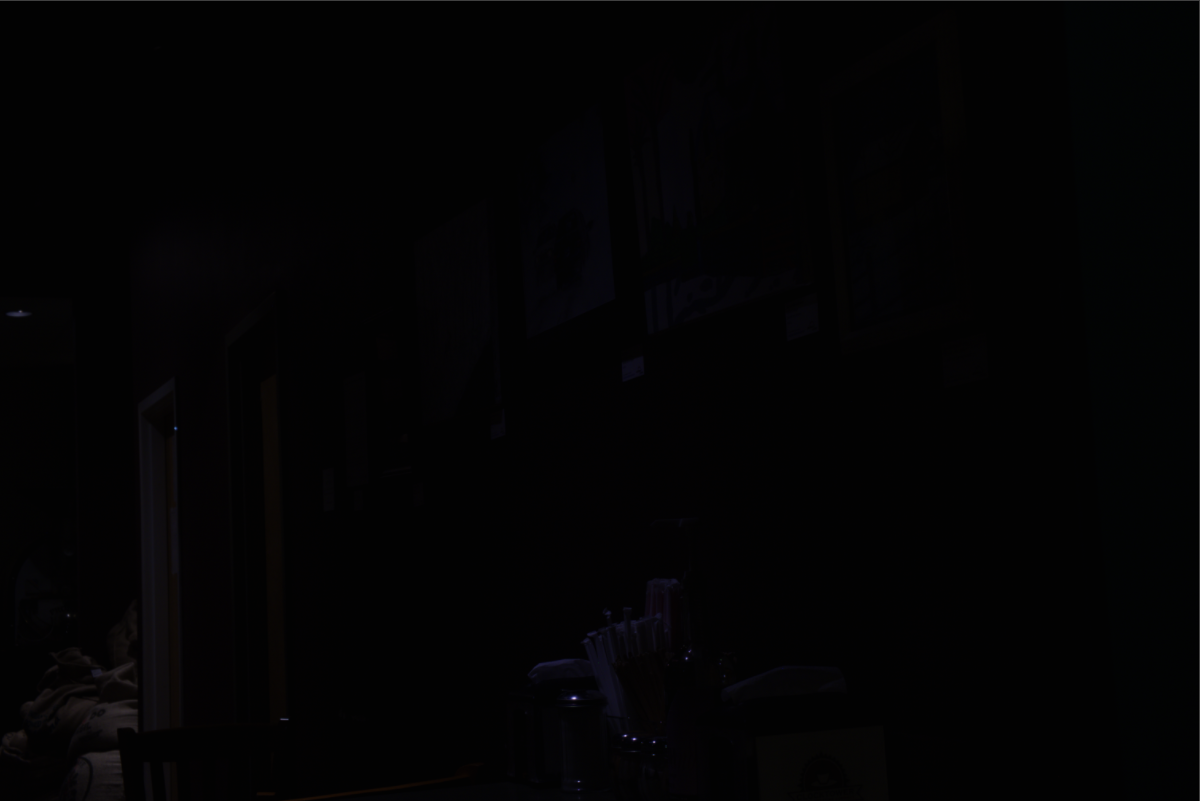}&
\includegraphics[width=0.2\linewidth]{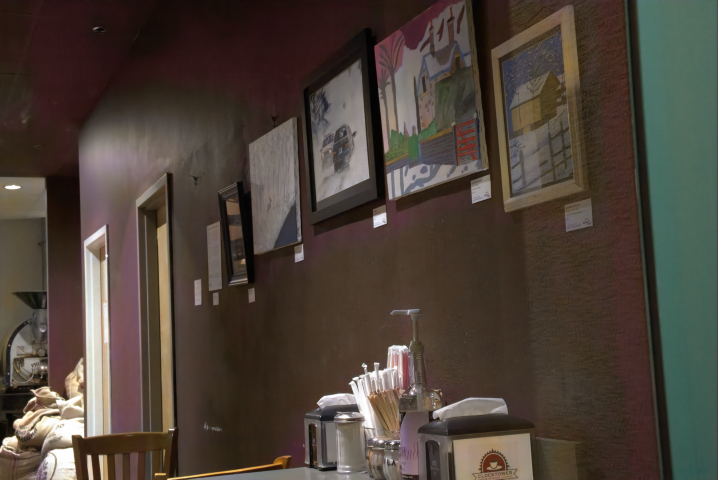}&
\includegraphics[width=0.2\linewidth]{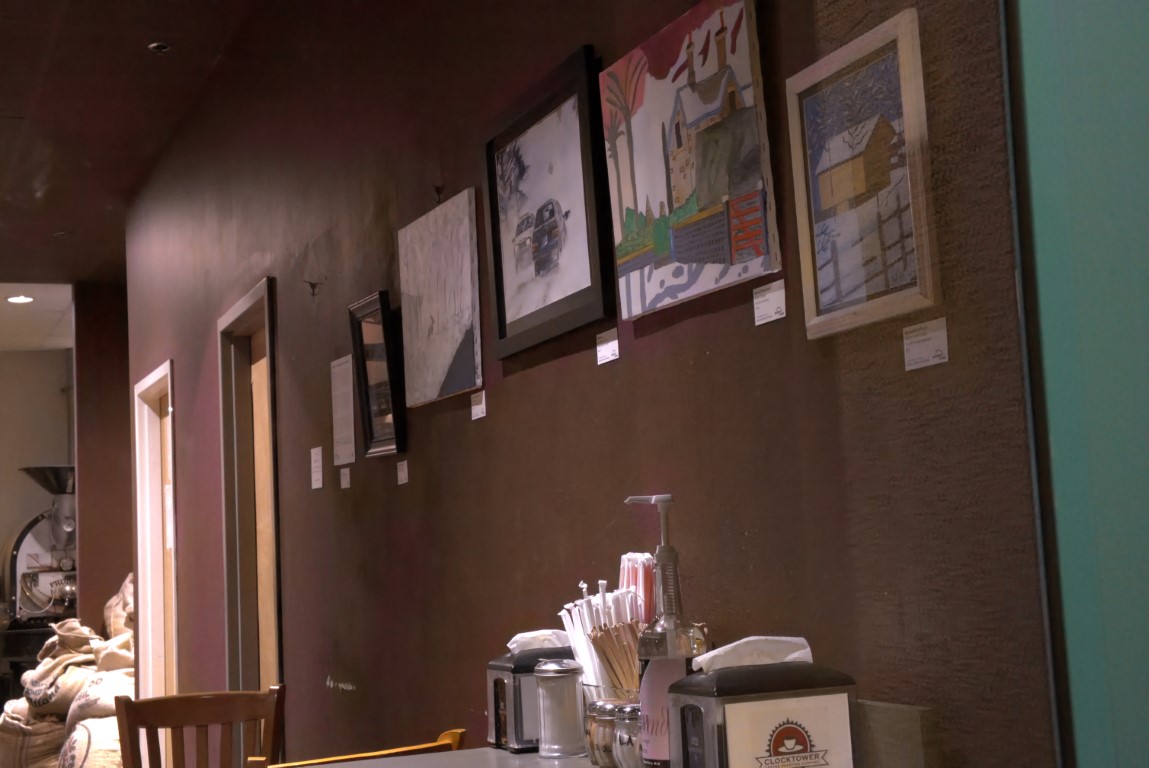}&
\includegraphics[width=0.2\linewidth]{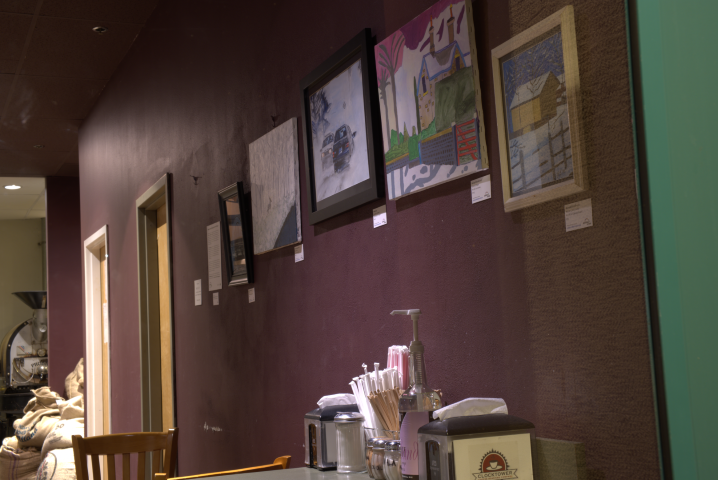}\\
\end{tabular}
\end{center}
\caption{Qualitative comparisons between DeepLPF and various state-of-the-art methods. See text for further details.}
\label{fig:visual_comparison}
\end{figure*}

\noindent \textbf{Quantitative Comparison: }
We compare DeepLPF image quality on MIT-Adobe-5K-DPE with contemporary methods (Table~\ref{tab:results1}, lower). Our full architecture is able to outperform recent methods; 8RESBLK~\cite{zhu2017unpaired,liu2017unsupervised}
and the supervised state-of-the-art (Deep Photo Enhancer (DPE)~\cite{chen2018deep}) on each of the three considered metrics, whilst our method parameter count approaches half that of their model capacity. Our implicitly regularised model formulation of filters allows similar quality enhancement capabilities yet is represented in a simpler form.

Similarly DeepLPF outperforms DeepUPE~\cite{wang2019underexposed}
on our second benchmark; MIT-Adobe-5K-UPE (\ie using their dataset splits and pre-processing protocols) when considering PSNR and LPIPS and provides competitive SSIM performance, and improves upon on all other compared works across the PSNR, SSIM metrics (see Table~\ref{tab:results2}). 

Finally we examine performance on the Fuji portion of the challenging SID dataset. Results are presented in Table~\ref{tab:resultsSID} where it can be observed that DeepLPF is able to improve upon the U-Net method presented in~\cite{chen2018learning} across all three considered metrics. Our method again proves more frugal, with model capacity lower by a factor of nearly four.

\begin{table}[t]
\centering
\caption{Quantitative comparison with state-of-the art methods on the \textbf{MIT-Adobe-5K-UPE} benchmark. PSNR and SSIM reported by competing works, are replicated from~\cite{chen2018deep}.}
\begin{adjustbox}{max width=0.495\textwidth}
\begin{tabular}{ p{34mm} | c c c c}
\multicolumn{5}{c}{} \\
\hline
\textbf{Architecture} & \textbf{PSNR}$\uparrow$  & \textbf{SSIM}$\uparrow$  & \textbf{LPIPS}$\downarrow$  & \textbf{\# Weights} \\
\hline
%DeepLPF           &${\textbf{24.39}}$ & ${0.885}$ & \textbf{0.107} & 1.8 M \\
DeepLPF &${\textbf{24.48}}$ & ${0.887}$ & \textbf{0.103} & \textbf{800K} \\
\hline
\hline
U-Net~\cite{Ronneberger15}                 & 22.24     & 0.850           & --    & 1.3 M \\
HDRNet~\cite{gharbi2017deep}               & 21.96     & 0.866           & --    & --    \\
DPE~\cite{chen2018deep}                    & 22.15     & 0.850           & --    & 3.3 M \\
White-Box~\cite{hu2018exposure}            & 18.57     & 0.701           & --    & --    \\
Distort-and-Recover~\cite{park2018distort} & 20.97     & 0.841           & --    & --    \\ 
DeepUPE~\cite{wang2019underexposed}        & 23.04     & \textbf{0.893}  & 0.158 & 1.0 M \\
\hline
\end{tabular}
\end{adjustbox}
\label{tab:results2}
\end{table}

\begin{table}[t]
\small
\centering
\caption{Quantitative performance comparisons for the image enhancement task defined by the RAW to RGB image pairs in the \textbf{SID} dataset (Fuji camera)~\cite{chen2018learning}.}
\begin{adjustbox}{max width=0.495\textwidth}
\begin{tabular}{ p{22mm} | c c c c }
\multicolumn{4}{c}{} \\
\hline
\textbf{\small{Architecture}} & \textbf{\small{PSNR}}$\uparrow$ & \textbf{\small{SSIM}}$\uparrow$   & \textbf{LPIPS}$\downarrow$ & \textbf{\# Weights}  \\
\hline
\small{DeepLPF}  & \textbf{26.82}  & \textbf{0.702} & \textbf{0.564} & \textbf{2.0 M}\\ 
%\small{DeepLPF (large)}  & \textbf{}  & \textbf{} & \sean{} & \textbf{5.0 M}\\ 
\hline
\hline
%\small{CAN}~\cite{chen2018learning}     & 25.71 & 0.710 & -- &  %-- \\ 
\small{U-Net}~\cite{chen2018learning}   & 26.61 & 0.680 &  0.586 &7.8 M \\ 
\hline
\end{tabular}
\end{adjustbox}
\label{tab:resultsSID}
\end{table}

\noindent \textbf{Qualitative Comparison:} Sample visual results comparing DeepLPF (trained independently using each of the three datasets) are shown in comparison to DPE~\cite{chen2018deep}, DeepUPE~\cite{wang2019underexposed} and SID (U-Net)~\cite{chen2018learning} models in Figure~\ref{fig:visual_comparison} rows, respectively. The park scene (first row) can be seen to visually improve with regard to reproduction of colour faithfulness in comparison to the DPE result. In the second row DeepUPE has overexposed the scene, whereas DeepLPF maintains both accurate exposure and colour content. The final row compares results on the challenging low-light dataset. The SID model output suffers from a purple colour cast whereas the DeepLPF output provides improved colour constancy in comparison to the ground truth. Finally, Figure~\ref{fig:filter_comparison1} provides a selection of parametric filters, represented by heatmaps, learned by our model. 
We provide additional qualitative results in the supplementary material.

\section{Conclusion}
\label{sec:conclusion}

In this paper, we have explored automated parameterisation of filters for spatially localised image enhancement. Inspired by professional image editing tools and software, our method estimates a sequence of image edits using graduated, elliptical and polynomial filters whose parameters can be regressed directly from convolutional features provided by a backbone network~\eg~U-Net. Our localised filters produce interpretable image adjustments with visually pleasing results and filters constitute plugable and reusable network blocks capable of improving image visual quality. 

In future work we can further explore automatic estimation of the optimal \emph{sequence} of filter application; \eg the Gumbel softmax trick~\cite{jang2017gumbel} may prove useful to select operations from a potentially large bank of image editing tools. We think that combining our presented local filters with additional local or global filter types and segmentation masks, refining enhancements to semantically related pixels, also provide interesting future directions towards interpretable and frugal automated image enhancement.

%\input{sec/supplementary.tex}

%\clearpage 
{\small
\bibliographystyle{ieee_fullname}
\bibliography{egpaper_for_review}
%\bibliography{supplementarybib}
}

\clearpage

\end{document}